\documentclass[letterpaper, 10 pt, conference]{IEEEtran}
\IEEEoverridecommandlockouts
\usepackage{cite}
\usepackage{amsmath,amssymb,amsfonts}
\usepackage{algorithmic}
\usepackage{graphicx}
\usepackage{textcomp}
\usepackage{xcolor}
\usepackage{comment}
\usepackage{balance}

\def\BibTeX{{\rm B\kern-.05em{\sc i\kern-.025em b}\kern-.08em
    T\kern-.1667em\lower.7ex\hbox{E}\kern-.125emX}}
\begin{document}

\title{A tool for emulating neuromorphic architectures with
memristive models and devices\\}

\author{Jinqi Huang, Spyros Stathopoulos, Alex Serb, and Themis Prodromakis\\
Email: \{j.huang, s.stathopoulos, a.serb, t.prodromakis\}@soton.ac.uk\\
Centre for Electronics Frontiers, Electronics and Computer Science, University of Southampton, UK\\
}

\maketitle

\begin{abstract}
Memristors have shown promising features for enhancing neuromorphic computing concepts and AI hardware accelerators. In this paper, we present a user-friendly software infrastructure that allows emulating a wide range of neuromorphic architectures with memristor models. This tool empowers studies that exploit memristors for online learning and online classification tasks, predicting memristor resistive state changes during the training process. The versatility of the tool is showcased through the capability for users to customise parameters in the employed memristor and neuronal models as well as the employed learning rules. This further allows users to validate concepts and their sensitivity across a wide range of parameters. We demonstrate the use of the tool via an MNIST classification task. Finally, we show how this tool can also be used to emulate the concepts under study in-silico with practical memristive devices via appropriate interfacing with commercially available characterisation tools.
\end{abstract}

\begin{IEEEkeywords}
 memristor, neuro-inspired computing, neuromorphic computing, neural networks, online learning, online classification, neuromorphic emulator, large scale characterisation
\end{IEEEkeywords}

\section{Introduction}

With recent interests in AI-related research, neuromorphic designs draw much attention because of their biological plausibility and high computational efficiency. Inspired by biology, neuromorphic designs benefit from efficient spike-based computation, data sparsity, and temporal coding. Dedicated neuromorphic hardware designs such as \cite{loihi, truenorth} overcome Von Neumann bottleneck by applying an in-memory computing design scheme, further allowing the ability to handle a large amount of data. Meanwhile, memristors \cite{memristor}, as a type of non-volatile devices, have been rapidly developed and have shown promising features for enhancing neuromorphic computing: extreme downscaling \cite{downscale}, high density integration \cite{high-density}, low switching energy with high switching rate \cite{high-speed-low-energy}, multi-bit storage \cite{multibit}, and physical properties similar to biological synapses \cite{stdp}. Memristors have become a popular candidate replacing SRAMs in in-memory computing -based dedicated neuromorphic hardware by storing information as resistive states (RSs). By directly applying Ohm's Law and Kirchhoff's Current Law, the current from the wordline side of a memristor array can be regarded as the dot product of two vectors which are represented as input voltages applied to the bitline side and memristor conductance respectively. As examples, ref \cite{bayesian} showed the potential of employing memristor arrays in Bayesian inference, and ref \cite{dot-product} presented memristor crossbar serving as a dot-product engine for accelerating neural networks. Ref \cite{insitu}, \cite{fully-hardware}, \cite{in-memory}, \cite{lstm}, and \cite{equi-accu} further performed matrix multiplication acceleration using memristor crossbar arrays in feedforward neural networks, convolutional neural networks, and long short-term memory. 

To further exploit the potential of memristors for enhancing neuromorphic systems, a dedicated simulation tool for fast validation of concepts and prediction of device states is needed. However, existing simulators such as MNSIM \cite{MNSIM} and NeuroSim \cite{neurosim} focus more on circuit-level behaviours, aiming at providing performance estimation in hardware. The simulator targeting general simulation of the system architecture with memristor models is still missing.

In this paper, we present a user-friendly software-based tool that enables simulations of neuromorphic computing with a wide range of neural network architectures, neuron models, learning rules and memristor settings. This tool allows users to validate the concept of exploiting memristors in neuromorphic systems, and to explore the parameter sensitivity of the system. Employed neuron models, learning rules and memristor models further allow users to customise their neuromorphic architectures by simply tuning parameters. A built-in analysis tool will also assist users to visualise inference results and internal variables for fast validation and debugging. Finally, this tool can also interface with commercially available characterisation tools such as \cite{arcone} to emulate designs with real devices.    

\section{Design Architecture}
\begin{figure*}[t]
    \centering
    \includegraphics[width=0.91\linewidth]{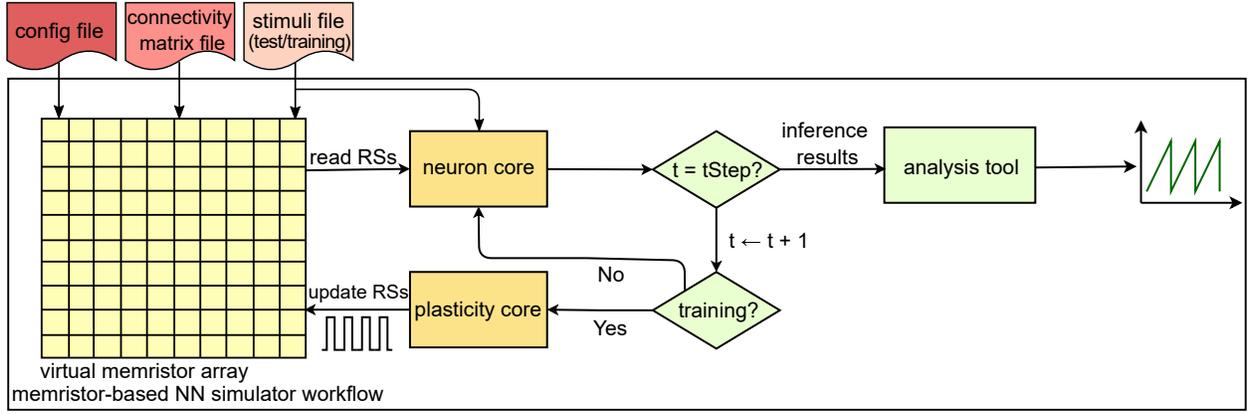}
    \caption{System workflow. Icons in different colors indicate different files in the system.}
    \label{fig:neuropack}
\end{figure*}

\subsection{Design overview}

Our work is a python-based software simulator aiming at performing simulation for neuromorphic architectures with memristor models. There are memristor models, neuron models, and learning rules embedded in the tool so that the simulation of the whole system performing real machine learning tasks can be easily set up by providing input data and system configuration parameters. Figure \ref{fig:neuropack} shows the workflow of this work. There are three independent input files: \textbf{configuration file}, \textbf{connectivity matrix file}, and \textbf{stimuli file}.
The \textbf{configuration file} provides parameters at both system level (including network size, network depth, and neuron numbers in each layer), and device level (such as memristor fitting parameters), to configure a neuromorphic architecture. The \textbf{connectivity matrix file} defines the connectivity of neurons and assigns memristors as synapses between neurons. The \textbf{stimuli file} stores incoming spikes fed to input neurons at each time step. After all input files are loaded to the tool, memristors models in 'virtual memristor array' are activated and initialised. Meanwhile, input spikes are decoded from the \textbf{stimuli file} and sent to the neuron model placed in 'neuron core'. In a single inference, 'neuron core' reads memristor RSs from 'virtual memristor array', and updates membrane voltages according to the specific neuron model selected by users. Fire history is then appended according to the determined threshold. If training is enabled, memristor RSs are also updated by the chosen learning rule in 'plasticity core' before starting another inference. The same process will repeat until all epochs have finished. Updated membrane voltages, weights and fire history can be loaded to a built-in analysis tool for further visualisation and debugging. 

There are two usage scenarios for this tool: one is to explore the sensitivity of the memristor-base neuromorphic system to a specific parameter. This tool allows users to monitor how memristor RS changes given different values of the target parameter. Another usage scenario is to validate algorithms, including neuron models and learning rules. Users can investigate a wide range of neuron models and learning, by selecting an example 'neuron core'/'plasticity core' or writing their own model using example files as templates. In both scenarios, the tool provides users options to customise their own neuromorphic architectures at both system level and device level.

\subsection{Neuron models and learning rules}

There are a wide range of neuron models and learning rules provided by this tool, including leaky integrate-and-fire (LIF) \cite{lif} and Izhikevich neuron \cite{Izhikevich} for neuron models and spiking-timing-dependent-plasticity (STDP) \cite{stdp}, back-propagation (BP) \cite{bp}, tempotron \cite{tempotron}, and direct random target projection (DRTP) \cite{drtp} for learning rules. Those options cover different application scenarios and design choices, from unsupervised learning to supervised learning, from rate-coding temporal coding, and from two-layer to multi-layer. This tool also allows users to explore their algorithms by simply using examples as templates. 

The main workflow of neuron model and learning rule is as follow: the neuron model placed in the 'neuron core' updates membrane voltages and determines neurons firing states, and the firing history is then sent to the learning rule placed in the 'plasticity core' to update weights if training is enabled. Once the condition of weight updating is met, pulses are sent to the 'virtual memristor array' to trigger weight updating. All actions are taken within one time step, and the same process will be repeated until the time step is equal to pre-set trail number. 

\begin{figure*}[t]
    \centering
    \includegraphics[width=0.9\linewidth]{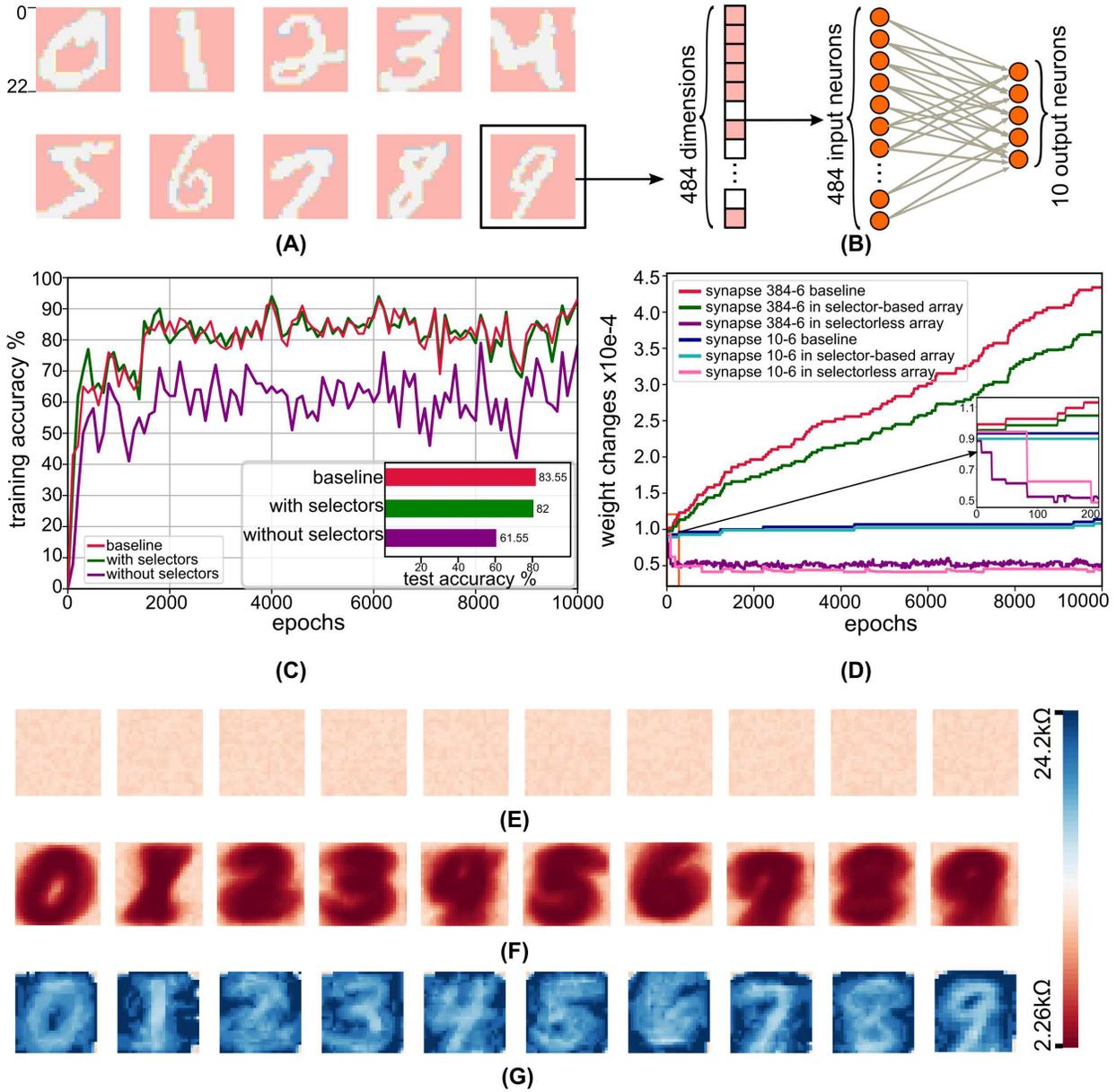}
    \caption{Simulation for handwritten digits recognition performed in selector-based and selectorless memristor arrays. (A) Input images. (B) Neural network architecture. (C) Accuracy curves. (D) Weight update curves. (E) Memristor resistive states before training, (F) after training in selector-based array, and (G) in selectorless array.}
    \label{fig:mnist}
\end{figure*}
\subsection{Memristor models}
Memristor models embedded in the tool are essential to provide a precise prediction for RSs given different triggering conditions. Here we use one empirical model \cite{memristor-model} which reflects the main switching characteristics of real memristor devices as an example, but this tool can be compatible with other user-defined memristor models as well to enable more flexibility. This memristor model can ideally emulate any type of device provided different values of parameters. The model explained how memristor resistance switching rate is dependent on bias voltages and current RSs. Model equations can be found below:

\begin{multline}\label{eq:switching}
    \frac{dR}{dt} = m(R, v) \\
    = 
     \begin{cases}
    A_p(-1+e^{\frac{|v|}{t_p}})(r_p(v) - R)^2& \text{if \( v >0,R<r_p(v)\)}  \\
    A_n(-1+e^{\frac{|v|}{t_n}})(R - r_n(v))^2& \text{if \(v \leq 0,R\geq r_n(v)\)}  
 \end{cases}
\end{multline} 

Where term \(A_{p, n}(-1+e^{\frac{|v|}{t_{p, n}}})\) explains the exponential dependency of the switching rate on bias voltage with \(A_{p, n}\) and \(t_{p, n}\) as scaling factors and fitting parameters respectively. Term \((r_{p, n}(v) - R)^2\) gives a fitting of second-order dependency of switching rate on current resistive states. Internal variables \(r_{p, n}\) represent resistive operating range given a bias voltage \(v\) with a first-order fitting function using fitting parameters \(a_{0p}\), \(a_{1p}\), \(a_{0n}\), \(a_{1n}\):

\begin{equation*}
    r(v) = 
    \begin{cases}
    r_p(v) = a_{0p}+a_{1p}v & \text{if \(v > 0\)}\\
    r_n(v) = a_{0n}+a_{1n}v & \text{if \(v\leq 0\)}
    \end{cases}
\end{equation*}

When a classification task is executed, 'virtual memristor array' takes user-defined parameters and initialises the memristor model. When a pulse is sent from the 'plasticity core' during the training phase, a resulting resistance is updated according to the model expressions and is returned to 'neuron core' when starting a read operation.

\subsection{Weights updating scheme} 
After an expected \(\Delta W\) is calculated in 'plasticity core' during the training phase, there is a critical step to write new weights to memristors. The weight update scheme used in this tool is to use 'predict-write-verify' loops until memristors resistive states converge to expected values. Firstly, when \(\Delta W\) is attained, the new expected RS is calculated based on the current RS and the weight mapping scheme. The resulting resistance is then calculated given options of pulse parameter sets provided by users. Next, set the pulse with parameters that lead to memristor RS closest to the expected before starting a read operation to verify. Finally, repeat the same process until the new resistance is within the R tolerance chosen by users, or the number of updating loops hits the maximum steps provided by users. The R tolerance is calculated as \((R_{expected} - R_{real})/R_{expected}\) to make sure the final memristor RSs are within an acceptable range, and the maximum step number is used to prevent the infinite loop if memristor RSs never converge to steady values due to the read noise.

\section{Application Example}

In this section, an application example using this tool to simulate handwritten digit recognition with MNIST dataset in both selector-based and selectorless memristor arrays will be presented and discussed. We used memristor parameters extracted from TiOx-based devices \cite{multibit} using extraction method proposed by \cite{extract}. We simulated with bias voltages ranging from \(\pm\)0.9V to \(\pm\)1.2V to predict how real devices behave in a real-world classification task. We are presenting two biasing scenarios: a) a scenario for selector based devices where bias is only applied to the selected device and b) a scenario with half-voltage mitigation for unselected devices to represent the selectorless case. In this task, original 28 \(\times\) 28 grey-scale images are cropped to 22 \(\times\) 22 and converted to binary format (see Figure \ref{fig:mnist} (A), pink and light grey pixels represent '1' and '0' respectively). The neural network used in this task is a 484 \(\times\) 10 winner-take-all network with leaky integrate-and-fire neuron model and gradient descent learning rule, as it is shown in Figure \ref{fig:mnist} (B). We map weights as device conductance without normalisation as most memristor-based neuromorphic designs do. Some key parameters used in this task are listed in Table \ref{tab:params}. To start the task, the memristor arrays are initialised to high resistance (low weights). In the ideal situation, weights in stimulated synapses will increase weight gradually, while those in non-stimulated synapses will stay high. Input images are firstly unrolled to 484-dimensional vectors whose bits are sent to each input neuron as 0-or-1 input spikes. If more than one output neurons are above the threshold, the one with the highest membrane voltage will fire while others will be inhibited. The network is firstly trained for 10000 epochs with a minibatch size of 100. Next, 2000 test data from a separated test set are sent to the network. We did tests for selector-based memristor array and selectorless memristor array as well as a version storing weights directly in the computer memory instead of memristors as the baseline. The accuracy curves can be found in Figure \ref{fig:mnist} (C). As it can be seen in the training accuracy curves, the general training accuracy of the baseline is 83.55\%, and selector-based array and selectorless array achieved 82\% and 61.55\% test accuracy respectively. This indicates two things: firstly the accuracy of selector-based array is mainly limited by factors out of memristor devices; secondly, there is a 20\% accuracy gap between selectorless and selector-based versions. 

To further explore what made the 20\% accuracy gap, the weight change curves for synapses connected between stimulated or non-stimulated input neuron and the target output neuron for both versions is displaced in Figure \ref{fig:mnist} (D), where synapse 384-6 represents a stimulated synapse and 10-6 non-stimulated. A stimulated synapse in the selector-based version (the dark green line) shows an increasing tendency during the whole training process, while in selectorless version (the dark purple line) weight was unexpectedly decreased when no update should occur (see the small window for the amplified region between epoch 0 and 200). This is due to unexpected weight updates in inactive devices caused by half bias voltages when target devices in the same bitlines or wordlines are inhibited. The impact of unwanted decrease is accumulated as many devices share the same bitlines or wordlines. For non-stimulated synapse (10-6), weight changes in selector-based version are trivial, while in selectorless version the synapse weight also decreased due to half bias voltages. However, even in the selectorless version, the weights in stimulated synapses are still slightly higher than those in the non-stimulated. The small weight differences still make the network able to learn images and be able to distinguish images in some cases. Figure \ref{fig:mnist} (E) - (G) show memristor resistive states before (E) and after training for both selector-based (F) and selectorless (G) versions. The small weight differences in the selectorless version are not distinguishable in a colourmap spanning a wide range, therefore we display resistive states instead. In Figure \ref{fig:mnist} array is initialised to \(\sim \)11k, and selector-based array has memristors with high resistance (\(\sim \)11k) in the background and with low resistance (\(\sim \)2.2k) in the middle to learn the digits. In the selectorless version, resistive states of memristors in the background have reached very high values (\(\sim \)24k), while those in the middle also have very high values (\(\sim \)20k) but are slightly smaller than those in the background.  

\begin{table}[t]
    \centering
    \caption{Key parameters used in the application example}
    \begin{tabular}{|c|c|}
    \hline
    Parameters & Values \\
    \hline
    array size & 100\(\times\)100 \\
    threshold &  10 mV\\
    learning rate & \(3.5 \times 10^{-6}\)\\
    \(A_p\) & 0.21389\\
    \(A_n\) & -0.81302\\
    \(t_p\) & 1.6591\\
    \(t_n\) & 1.5148\\
    \(a_{0p}\) & 37087\\
    \(a_{0n}\) & 43430\\
    \(a_{1p}\) & -20193\\
    \(a_{1n}\) & 34333\\
    R tolerance & 0.1\%\\
    max update steps & 5\\
    \hline
    \end{tabular}
    \label{tab:params}
\end{table}

\section{Discussion} 
In this paper, we show a software-based tool for emulating memristor-based neuromorphic architectures. This tool can work as a standalone simulator to allow users to explore different neuron models, learning rules, memristor models, different numbers or types of memristor devices, different neural network architectures, and different neuromorphic applications by simply providing different values of parameters or interfacing user-defined models with the tool. This tool can also be connected to commercially available memristor characterisation instruments to further emulate designs with real memristor devices. As an example application, we show simulation results of handwritten digit recognition in both selector-based and selectorless memristor arrays, and we further explore how and why the performance gap exists. 



\balance

\vspace{12pt}

\end{document}